# Improving the ROS 2 Navigation Stack with Real-Time Local Costmap Updates for Agricultural Applications

Ettore Sani   Antonio Sgorbissa   Stefano Carpin

*Abstract*— The ROS 2 Navigation Stack (Nav2) has emerged as a widely used software component providing the underlying basis to develop a variety of high-level functionalities. However, when used in outdoor environments such as orchards and vineyards, its functionality is notably limited by the presence of obstacles and/or situations not commonly found in indoor settings. One such example is given by tall grass and weeds that can be safely traversed by a robot, but that can be perceived as obstacles by LiDAR sensors, and then force the robot to take longer paths to avoid them, or abort navigation altogether. To overcome these limitations, domain specific extensions must be developed and integrated into the software pipeline. This paper presents a new, lightweight approach to address this challenge and improve outdoor robot navigation. Leveraging the multi-scale nature of the costmaps supporting Nav2, we developed a system that using a depth camera performs pixel level classification on the images, and in real time injects corrections into the local cost map, thus enabling the robot to traverse areas that would otherwise be avoided by the Nav2. Our approach has been implemented and validated on a Clearpath Husky and we demonstrate that with this extension the robot is able to perform navigation tasks that would be otherwise not practical with the standard components.

## I. INTRODUCTION

Outdoor robot navigation presents numerous challenges to autonomous robots due to the unpredictability of natural environments. In agriculture, where the terrain is often uneven, vegetation is diverse and uncontrolled, and lighting conditions can vary significantly, the demand for robust autonomous robots is pressing. For robots operating in these outdoor settings, the ability to make real-time decisions informed by sensorial perceptions is essential.

One of the central difficulties in outdoor robot navigation, especially in agricultural settings, is the effective recognition of traversable space when moving in environments with tall grass and/or small bushes. Figure 1 illustrates such a situation. Off-the-shelf obstacle detection systems tuned for indoor or structured environments often categorizes these traversable obstacles as permanent ones, resulting in sub-optimal path planning, or even compromising the planning process altogether.

E. Sani and A. Sgorbissa are with the Department of Informatics, Bioengineering, Robotics, and Systems Engineering, University of Genova. S. Carpin is with the Department of Computer Science and Engineering, University of California, Merced, CA, USA. S. Carpin is partially supported by USDA-NIFA under award # 2021-67022-33452 (National Robotics Initiative) and by the IoT4Ag Engineering Research Center funded by the National Science Foundation (NSF) under NSF Cooperative Agreement Number EEC-1941529. Any opinions, findings, conclusions, or recommendations expressed in this publication are those of the author(s) and do not necessarily reflect the view of the U.S. Department of Agriculture or the National Science Foundation.

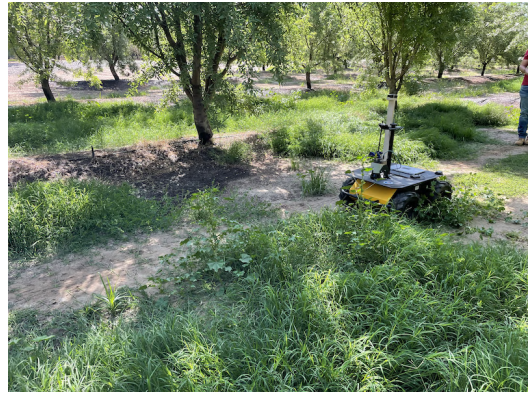

Fig. 1: A robot navigating in an almond orchard facing high weeds and bushes that are classified as obstacles by the standard ROS 2 navigation stack, but that can be safely traversed.

Addressing the challenge of recognizing traversable obstacles in outdoor robotics navigation has significant implications for the practical deployment of autonomous systems to solve real-world tasks. More accurate detection and differentiation of traversable areas can lead to more efficient path planning, reducing the risk of getting stuck or lost in challenging outdoor environments.

ROS and ROS 2 have emerged as the de-facto standard for numerous robotic applications and the navigation stack Nav2 [10] has proved to be an essential building block to develop more high-level capabilities, such as data collection, anomaly detection and the like. However, Nav2 is mostly tailored towards indoor applications and heavily relies on input coming from sensors such as LiDARS that treat weeds and bushes as untraversable space, thus making this software module not fully adequate for field operations.

In this context, we introduce a new approach to robustify outdoor robot navigation relying on Nav2. Our primary objective is to develop a comprehensive solution for enhancing outdoor robot navigation that can be transparently integrated into Nav2 and enable the robot to navigate through areas with high weeds or bushes that can be safely traversed. Specifically, we implemented a lightweight real-time computer vision system that accurately identifies traversable areas within the robot's field of view, even in the presence of environmental features classified as non-traversable by the LiDAR, and then injects its output into the ROS local costmap, allowing for dynamic path planning and navigation in real-world outdoor environments. Exploiting the *plugin* architecture used to implement the Nav2, our solution can be seamlessly integrated into existing systems without compromising other functionalities, and, importantly, without the need to modify or reconfigure any other software component inside or outside Nav2. Our contribution does not modify Nav2; rather, it introduces a smart solution that utilizes the ROS Layered Costmap to incorporate custom information from the robot's vision system. This strategy ensures that our method can be seamlessly adopted by the agricultural robotics community, which might resist modifications to the standard ROS navigation framework.

The rest of this paper is organized as follows: Section II reviews the literature on outdoor robot navigation and discusses relevant approaches. Our technical approach is presented in Section III where we detail the technical aspects of our solution, including sensor integration, real-time image segmentation using YOLOv8, dataset preparation, and integration with Nav2. Section IV presents the results of our experiments, including performance metrics and real-world applications of our system with on-the-field experiments conducted in agricultural orchards. Finally, in Section V we summarize the findings and the contributions of our research and discusses future directions.

## II. RELATED WORK

Outdoor robot navigation remains a complex and evolving field, with numerous challenges posed by unstructured environments. This section provides an overview of recent developments and relevant research, showcasing the contributions of this paper in this area. Earlier solutions to solve similar problems are presented in [13], [21], [22].

The work of Guan et al. [4] uses computer vision for terrain segmentation to improve robot navigation in outdoor environments. While our approach aligns with segmenting images to identify traversable areas, it is different in that it focuses on enhancing path planning where the laser sensor detects overcomable obstacles. This approach utilizes computer vision techniques to classify terrain in the robot's field of view based on navigability levels using RGB images. Reference [15] also shares a similar point of view. The approach of [19] instead segments the robot's surroundings through anomaly detection, identifying situations or terrains that fall outside the norm or training data. This approach can help the robot navigate safely in diverse environments by recognizing unfamiliar or potentially hazardous conditions.

While these approaches are intriguing, they are suited in scenarios where the terrain in front of the robot exhibits substantial variability. Nonetheless, the agricultural field of work is more likely to involve a robot operating within a lane or predefined path in an orchard. In such scenarios, despite the terrain's relatively consistent nature, the robot may face difficulty planning a path to its next goal due to the presence of overcomable obstacles detected by the laser sensor. Therefore, addressing these specific kinds of obstacles is of paramount relevance for agricultural robot navigation.

In our research, we drew inspiration from numerous prior work [6], [14], [17], [20] to construct our custom dataset for training a neural network for terrain classification. In particular Procopio et al. [14] provide a valuable contribution for autonomous robot navigation within unstructured environments, introducing an ensemble learning strategy that combines multiple classifiers to segment terrains.

The research presented in the article by Martini et al. [11] focuses on addressing navigation challenges in agricultural contexts, particularly vineyards and orchards, by developing a lightweight and flexible navigation solution that does not rely on precise localization data. One notable aspect of this approach is its reliance on depth images and position-agnostic robot state information. However, it is important to note that the experiments conducted in this work are primarily in simulated vineyard environments.

In outdoor robot navigation, Dima and Hebert [2] advanced supervised learning, particularly in obstacle detection, revealing active learning's potential for reducing labeling efforts.

The work by Weerakoon et al. [18] offers another approach to enhancing robot navigation in uneven outdoor terrains. Their method relies on leveraging deep reinforcement learning (DRL) to address the challenges of unpredictable and rugged environments, combining elevation maps, robot pose, and goal information as inputs for a fully-trained network. The approach presented in [8] also addresses the same problem with DRL.

Guan et al. [5] combine visual and inertial data to classify traversable surfaces and enhance robotic navigation. The authors developed an innovative navigation-based labeling scheme for terrain classification, employing a combination of RGB images and inertial measurement unit (IMU) data. This fusion of visual and inertial information enhances the ability to accurately classify traversable surfaces and adaptively guide robotic navigation across diverse terrains.

Lastly, the article by Silver et al. [16] addresses the task of autonomous navigation in heavily unstructured environments. The study emphasizes the crucial interplay between perception and planning systems and highlights their impact on the robot's behavior and overall performance.

## III. TECHNICAL APPROACH

In this section, we present the details of our proposed method for enhancing outdoor robot navigation through modified perception. Our method has been developed and tested on the robot shown in Figure 2. It is a Clearpath Husky robot tasked with navigating in pistachio and almond orchards to collect images of trees for early disease detection and growth monitoring. It includes a Sick LMS111 LiDAR in the front, a Bosch BNO085 9 axis IMU, a Ublox GNSS receiver, and an Oak-D 3D camera by Luxonis. The robot is controlled by an onboard laptop with Ubuntu 22 and ROS 2 Humble.

The algorithmic pipeline shown in Figure 3 unfolds as follows: integration of sensor inputs, use of YOLOv8 image segmentation algorithm, generation of a PointCloud2 representation, and integration with the Nav2 framework via a custom costmap plugin.

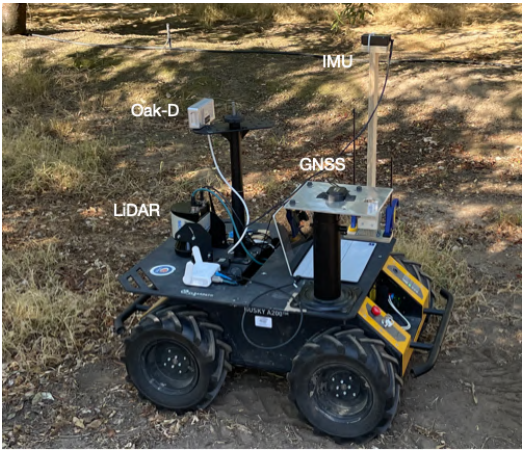

Fig. 2: Field experiment in orchard: The Clearpath Husky robot, equipped with sensors including a laser sensor, IMU, GNSS, and an Oak-D camera, navigating autonomously during an experiment.

*A. Sensor Inputs*

The input to our pipeline comes from the Oak-D camera. This device provides both RGB images and corrected depth images. The depth information derives from a stereo sensor configuration, where two images with a calibrated baseline separation are triangulated to yield accurate and reliable depth maps suitable for outdoor applications. Moreover, to ensure the fidelity of our data inputs, we performed synchronization procedures to guarantee the temporal alignment and dimensional congruence of the paired RGB and depth images. More precisely, for message synchronization, we used the ROS 2 *message_filter* package, a utility library for managing and synchronizing messages within ROS 2. It allows for the precise alignment of messages from various sensors based on their timestamps. We employed the approximate time synchronizer filter, which processes messages of varying types from multiple sources. This filter triggers the corresponding callback only when it receives messages from each of these sources with timestamps falling within a defined threshold of 0.1 seconds.

*B. Nav2 Integration*

Integration into Nav2 is achieved by generating a PointCloud2 representation of the traversable terrain at the end of our pipeline, as described in Section III-E. To merge this information with the Nav2 Local Costmap, a custom component typically used by the Nav2 planner for finding a path, we developed a Custom Costmap Plugin, serving as a component of this integrated system.

Nav2's Local Costmap continuously builds and updates a dynamic representation of the robot's surroundings, and it features different layers. Each layer contributes specific information to the Local Costmap by assigning varying cost values to different parts of the robot's local map, with higher costs denoting obstacles and lower costs representing open spaces. The layered structure combines these cost values from all layers to create a comprehensive map. Examples of these layers include:

- Static Layer: Assigns costs based on known, static obstacles like walls or trees.
- Obstacle Layer: Uses sensor data to detect dynamic obstacles such as pedestrians or moving objects.
- Voxel Layer: Utilizes voxel grid representations to handle complex three-dimensional environments.
- Inflation Layer: Inflates cost values around obstacles to ensure a safety navigation margin for the robot.

Nav2's default behavior is to populate the Local Costmap with obstacles identified by the laser sensor. As the robot moves, this costmap is continually updated to reflect real-time changes in its surroundings. The Local Costmap provides essential information for collision avoidance and path planning algorithms, allowing the robot make informed decisions and adjust its trajectory to reach its intended destination.

The plugin structure of Nav2 is designed to enhance its modularity and flexibility, and every costmap plugin constitutes a layer of the costmap. Accordingly, our computer vision algorithm is implemented as a plugin because it allows to inject its inputs into one of the layers without affecting the core Nav2 functionality. This flexibility ensures that our computer vision-based approach for terrain segmentation, which can improve path planning in some scenarios, can be easily integrated or excluded from the navigation pipeline as needed.

In real-world contexts it often happens that the laser scanner detects areas potentially traversable as obstacles, such as vegetation or taller grass, which might be erroneously categorized as zones to avoid the costmap. As this happens, the Custom Plugin integrates the PointCloud2 data derived from the computer vision algorithm, clearing all zones detected as traversable. Figure 4 shows a schema of the Nav2 layered costmap's behavior with our Custom Costmap Plugin. Placed after the Voxel Layer, our Custom Layer integrates the information of traversable zones by changing the Local Costmap. Consequently, this method enables the Nav2 framework to plan trajectories through areas initially identified as obstacles, thereby unveiling potentially shorter pathways toward the goal.

*C. YOLOv8 Image Segmentation*

We utilized the YOLOv8 [7] deep neural network-based object detection algorithm as an off-the-shelf solution to achieve accurate image segmentation. YOLOv8 operates on RGB images and generates a segmentation mask that partitions the image into distinct regions namely traversable space and obstacles. Subsequently, the regions identified as traversable are used as the basis of the fusion of this information into our Local Costmap. The YOLOv8 model was trained using a custom dataset comprising over 1000 manually labeled images.

The motivation for using YoloV8, which offers instance segmentation capabilities, as opposed to a solution that performs only semantic segmentation without identifying individual instances lies in its state-of-the-art performance and computational requirements. However, it is important to mention that for the specific issue addressed, any existing multiple instances and merged, and solely the semantic pixel-level information provided by YoloV8 is utilized. After areas without obstacles are labeled, determining whether an area is traversable also depends on the robot's physical characteristics, such as its clearance underneath. This issue is managed by the Nav2 Costmap Inflation layer, Figure 4.

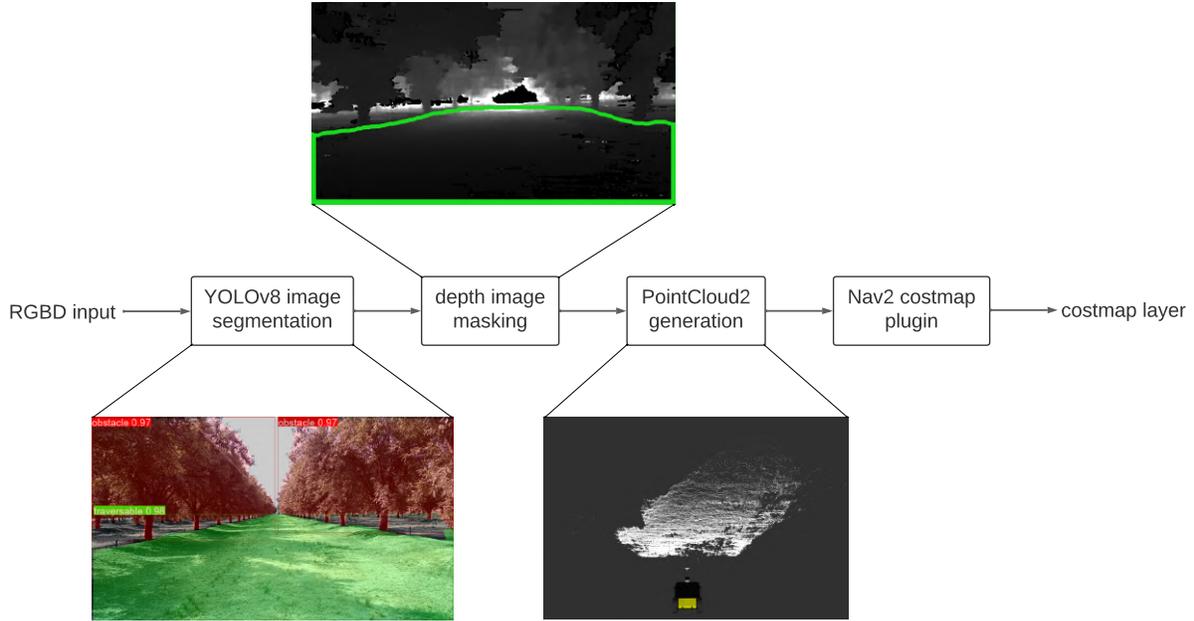

Fig. 3: Schema of the proposed pipeline: from RGBD images to Nav2 Costmap.

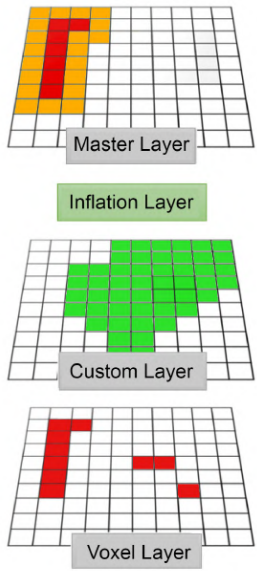

Fig. 4: Schema of the Nav2 Layered Costmap. The Custom Layer, positioned above the Voxel Layer, integrates information regarding traversable zones, thereby modifying the Local Costmap. The red pixels denote obstacles identified by the Voxel Layer, while the green pixels represent areas cleared by our custom plugin. The yellow pixels are generated by the Inflation layer, which expands obstacles to ensure safe navigation within the Nav2 stack.

### D. Custom Dataset

With meticulous labeling, we assembled a dataset that encompasses various outdoor scenarios for training the model to accurately identify and segment obstacles and traversable areas within the images. The dataset consists of 1066 annotated images with a considerable proportion, approximately 85%, drawn from outdoor experiments conducted as an integral part of this research endeavor. Figure 5 shows a sample of these images for reference. Supplementary images were sourced from publicly available outdoor datasets [12] [1] and we manually labeled both collected and external dataset images to ensure homogeneity. The annotation process, facilitated by the Roboflow software [3], categorized each image's regions into two primary classes, i.e., "traversable" and "obstacle" zones, while designating the remaining sections as "do not care" regions (see Figure 6 for an example). We used a range of augmentation techniques each targeted at introducing variance and complexities into the training data. These strategies, including image flipping, probabilistic grayscale conversion, saturation adjustment, brightness variation, Gaussian blurring, exposure fluctuations, and pixel-level noise addition, foster the model's adaptability and generalization capabilities. By applying these data augmentation techniques, we expanded the dataset's diversity and size, leading to better generalization and performance of the neural network in image segmentation tasks. The enriched dataset allowed the model to learn from more scenarios, resulting in more accurate predictions in challenging real-world environments.

### E. Point Cloud Generation

Building upon the segmentation mask obtained from YOLOv8, we apply this mask to the corresponding corrected depth image. For this process, we ensure that the two images, RGB and depth, have congruent dimensions and refer to the same time instant. This operation results in extracting depth information only from traversable areas, resulting in a 3D reconstruction of the zones that the robot can safely traverse.

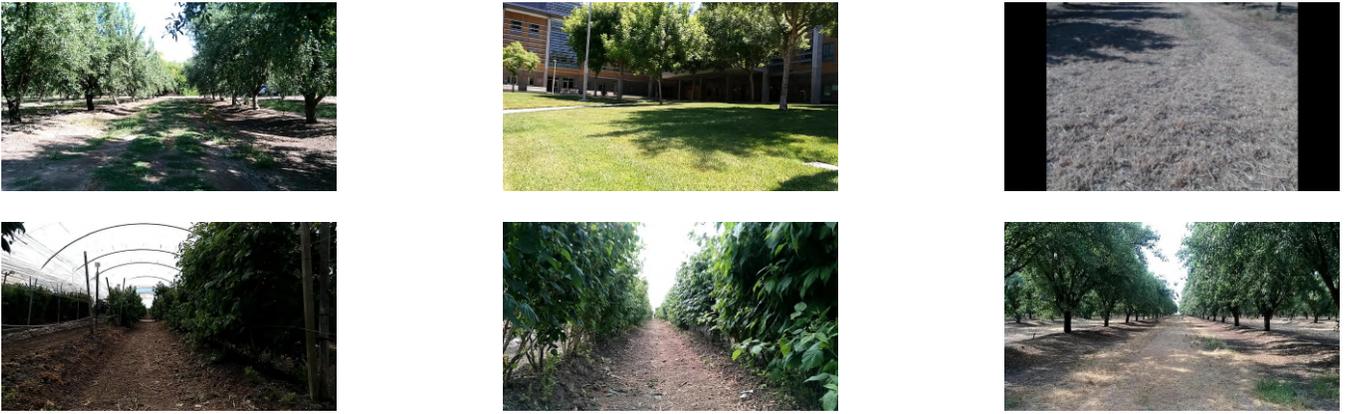

Fig. 5: Example of images in the dataset.

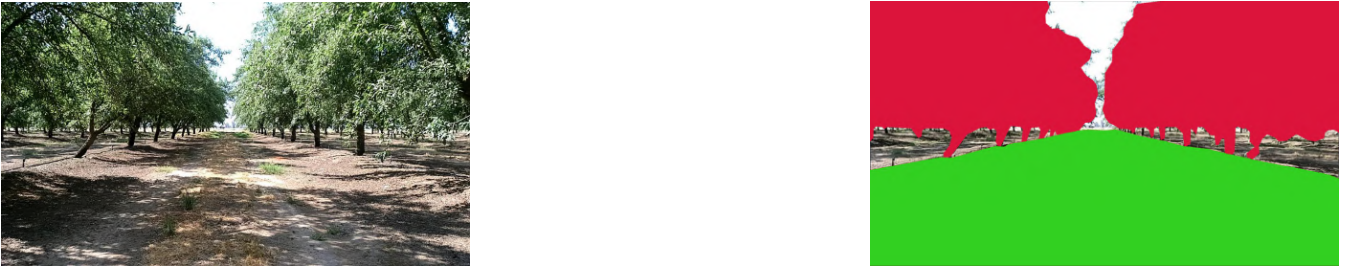

Fig. 6: Example of ground truth manual annotation in the dataset: (a) Original image and (b) Image with manually applied labels. Red pixels correspond to "obstacle", while green pixels correspond to "traversable".

To transform a segmented depth image to a point cloud, we utilize the ROS 2 *depth_image_proc_node*, that generates a PointCloud2 message as a structured and concise representation of the 3D environment. This message encapsulates spatial information about the regions the robot can safely navigate, as shown in Figure 7.

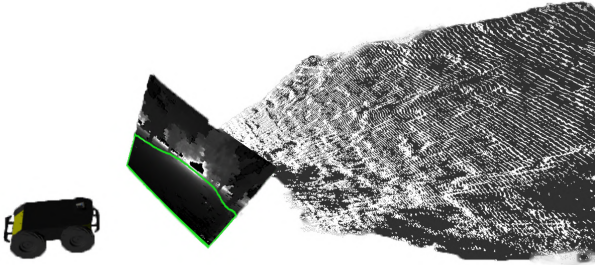

Fig. 7: Point Cloud of the traversable area generated from the segmented depth image.

## IV. EXPERIMENTAL EVALUATION

### A. Neural Network Training

We performed training on different YOLOv8 architectures, with the main objective of exploring the tradeoffs between accuracy and inference speed, given that the classification step at run time has to be executed frequently. The dataset was divided into three subsets: the Training set (70%), the Validation set (20%), and the Test set (10%). Evaluation results based on the Validation set are outlined in Table I. All models were trained on an NVIDIA GeForce RTX 3090 GPU within the PyTorch 2.0.1 environment. Furthermore, we leveraged YOLOv8's built-in RayTune [9] utility for hyperparameter optimization. It is important to note that this feature was exclusively compatible with the two smallest architectures, namely, yolov8n and yolov8s. We evaluated each trained network on the basis of mean Average Precision (mAP), inference speed, and model size. In particular, inference time holds critical significance due to its influence on the overall timing of the algorithm and power consumption. We directly evaluated the Average Inference Time with the entire algorithm running on an Intel i7 Core processor without GPU. Based on the evaluation of the trained neural networks, we opted for the yolov8n-seg model, utilizing RayTune hyperparameter optimization. This selection was motivated by its low inference time, which aligns with our stringent real-time requirements. Additionally, the model offered a reasonable mean Average Precision, making it a well-suited choice for our specific needs in outdoor robot navigation. The scores on the Test set are: 0.810 mAP@0.5, 0.632 mAP@0.5:0.95, 0.832 Precision, and 0.783 Recall.

### B. On-field test

Figure 8 illustrates the final outcome of the pipeline. By identifying and clearing traversable zones, the plugin sets specific sections of the map as free. More comprehensive results are featured in the video associated with this submission. Figure 8c visualizes an example of the outcomes of the computer vision process. Notably, on the right-hand side, the region of terrain containing tall grass is correctly categorized as traversable (green). The impact of our proposed pipeline becomes evident when comparing Figures 8a and 8b. In Figure 8a, the tall grass in front of the robot is detected as an obstacle by the laser sensor, leading to its inclusion in the standard Nav2 costmap as an obstacle.

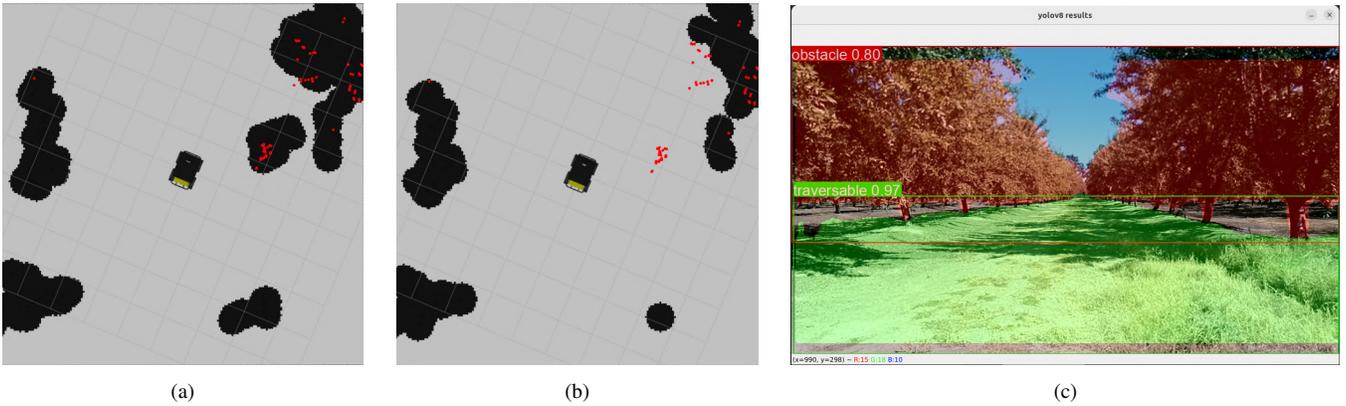

Fig. 8: Real-time behavior comparison of the custom plugin: (a) Costmap generated using standard Nav2 without the custom plugin; (b) Costmap generated with the custom plugin, which clears zones with grass detected as traversable; (c) Corresponding segmented image captured during this experiment.

| Model | Size | mAP@0.5 | mAP@0.5:0.95 | Precision | Recall | Average Inference Time |
|---|---|---|---|---|---|---|
| yolov8n-seg | 6.5 MB | 0.795 | 0.604 | 0.851 | 0.716 | 32.9 ms |
| yolov8s-seg | 23 MB | 0.805 | 0.617 | 0.843 | 0.748 | 77.0 ms |
| yolov8m-seg | 53 MB | 0.832 | 0.648 | 0.883 | 0.770 | 176.6 ms |
| yolov8l-seg | 89 MB | 0.840 | 0.655 | 0.863 | 0.800 | 336.0 ms |
| yolov8x-seg | 138 MB | 0.845 | 0.665 | 0.856 | 0.796 | 553.5 ms |
| **yolov8n-seg R** | **6.5 MB** | **0.822** | **0.631** | **0.827** | **0.773** | **32.9 ms** |
| yolov8s-seg R | 23 MB | 0.845 | 0.665 | 0.870 | 0.795 | 76.9 ms |

TABLE I: Comparison of YOLOv8 Architectures: Training Results on the Validation Set and Average Inference Times for real-world applications. (R: result obtained with RayTune to optimize hyperparameters.)

However, as shown in Figure 8b, the custom plugin efficiently removes this obstacle from the costmap, enabling the robot to plan paths with fewer obstructions, thus leading to shorter paths to its goal. The portion of terrain initially identified as an obstacle due to tall grass by the laser sensor is now recognized as free space. This snapshot was captured during an experiment conducted in an orchard, showcasing the real-time applicability of this solution. Our proposed method addresses the challenge of outdoor robot navigation through the modification of perception. By combining RGB and corrected depth images, utilizing YOLOv8 for image segmentation, generating PointCloud2 representations, and interfacing with the Nav2 framework, our pipeline enhances path planning performance in scenarios with traversable and real obstacles. Finally, we outline that our software is made freely available as open source on GitHub at https://tinyurl.com/ycyp6df8.

## V. CONCLUSIONS

In this work we proposed a method to enhance outdoor robot navigation, focusing on the challenges faced in real-world agricultural environments. Our goal was to create a comprehensive pipeline that could robustly detect traversable zones, overcoming limitations where areas covered in tall grass are treated as obstacles.

To accomplish our task, we created a custom dataset, collecting images from diverse outdoor scenarios and applying manual annotations. This dataset was the foundation for training our chosen neural network, YOLOv8, known for its real-time capabilities. The core of our approach lies in the fusion of pixel level visual classification with the Nav2 framework. By integrating real-time image segmentation into the robot's perception, we enable it to distinguish traversable areas from obstacles.

Our custom costmap plugin, designed to interface with the Nav2 framework, ensures that the information retrieved from computer vision is integrated into the robot's costmap, enabling the path planner to make more informed decisions regarding the terrain around the robot. Our evaluation process, including real-world experiments, underscores the effectiveness of our approach.

This work does not modify Nav2, ensuring that our method can be adopted by researchers who are already familiar with the ROS navigation framework. However, it introduces a smart solution that significantly expands the applicability of Nav2, while retaining its strengths. By augmenting the robot's perception with real-time image segmentation, we enhanced outdoor navigation, a crucial step in autonomous agricultural robotics. We think that our pipeline represents a valuable contribution to the field, addressing the challenges of real-world agricultural environments. Future work will include more systematic evaluation in the field.

## REFERENCES

[1] FLying Wedge defence and aerospace. tree detection model dataset. https://universe.roboflow.com/flying-wedge-defence-and-aerospace-dcr4s/tree-detection-model, aug 2023. visited on 2023-08-29.

[2] Cristian Dima and Martial Hebert. Active learning for outdoor obstacle detection. In *Robotics: Science and Systems*, pages 9–16, 2005.

[3] Nelson J. Dwyer B. Roboflow, 2023.

[4] T. Guan, D. Kothandaraman, R. Chandra, A.J. Sathyamoorthy, K. Weerakoon, and D. Manocha. Ga-nav: Efficient terrain segmentation for robot navigation in unstructured outdoor environments. *IEEE Robotics and Automation Letters*, 7(3):8138–8145, 2022.

[5] T. Guan, R. Song, Z. Ye, and L. Zhang. Vinet: Visual and inertial-based terrain classification and adaptive navigation over unknown terrain. In *Proceedings of the IEEE International Conference on Robotics and Automation*, pages 4106–4112, 2023.

[6] P. Jiang, P. Osteen, M. Wigness, and S. Saripalli. Rellis-3d dataset: Data, benchmarks and analysis. In *Proceedings of the IEEE International Conference on Robotics and Automation*, pages 1110–1116, 2021.

[7] Glenn Jocher, Ayush Chaurasia, and Jing Qiu. Ultralytics yolov8, 2023.

[8] S. Josef and A. Degani. Deep reinforcement learning for safe local planning of a ground vehicle in unknown rough terrain. *IEEE Robotics and Automation Letters*, 5(4):6748–6755, 2020.

[9] R. Liaw, E. Liang, R. Nishihara, P. Moritz, J.E. Gonzalez, and I. Stoica. Tune: A research platform for distributed model selection and training. *arXiv preprint arXiv:1807.05118*, 2018.

[10] S. Macenski, F. Martin, R. White, and J. Ginés Clavero. The marathon 2: A navigation system. In *Proceedings of the IEEE/RSJ International Conference on Intelligent Robots and Systems IEEE/RSJ International Conference on Intelligent Robots and Systems (IROS)*, pages 2718–2725, 2020.

[11] M. Martini, S. Cerrato, F. Salvetti, S. Angarano, and M. Chiaberge. Position-agnostic autonomous navigation in vineyards with deep reinforcement learning. In *Proceedings of the IEEE International Conference on Automation Science and Engineering*, pages 477–484, 2022.

[12] new-workspace mxtet. Fences dataset. https://universe.roboflow.com/new-workspace-mxtet/fences-maq01, dec 2021. visited on 2023-08-29.

[13] G.A.S. Pereira, L.C.A. Pimenta, A.R. Fonseca, L. de Q. Corrêa, R.C. Mesquita, L. Chaimowicz, D.S.C. de Almeida, and M.F.M. Campos. Robot navigation in multi-terrain outdoor environments. *The International Journal of Robotics Research*, 28(6):685–700, 2009.

[14] M.J. Procopio, J. Mulligan, and G. Grudic. Learning terrain segmentation with classifier ensembles for autonomous robot navigation in unstructured environments. *Journal of Field Robotics*, 26(2):145–175, 2009.

[15] A.J. Sathyamoorthy, K. Weerakoon, T. Guan, J. Liang, and D. Manocha. Terrapn: Unstructured terrain navigation using online self-supervised learning. In *Proceedings of the IEEE/RSJ International Conference on Intelligent Robots and Systems*, pages 7197–7204, 2022.

[16] D. Silver, J.A. Bagnell, and A. Stentz. Learning from demonstration for autonomous navigation in complex unstructured terrain. *The International Journal of Robotics Research*, 29(12):1565–1592, 2010.

[17] K. Viswanath, K. Singh, P. Jiang, P.B. Sujit, and S. Saripalli. Offseg: A semantic segmentation framework for off-road driving. In *Proceedings of the IEEE International Conference on Automation Science and Engineering*, pages 354–359, 2021.

[18] K. Weerakoon, A.J. Sathyamoorthy, U. Patel, and D. Manocha. Terp: Reliable planning in uneven outdoor environments using deep reinforcement learning. In *Proceedings of the IEEE International Conference on Robotics and Automation*, pages 9447–9453, 2022.

[19] L. Wellhausen, R. Ranftl, and M. Hutter. Safe robot navigation via multi-modal anomaly detection. *IEEE Robotics and Automation Letters*, 5(2):1326–1333, 2020.

[20] M. Wigness, S. Eum, J.G. Rogers, D. Han, and H. Kwon. A rugd dataset for autonomous navigation and visual perception in unstructured outdoor environments. In *Proceedings of the IEEE/RSJ International Conference on Intelligent Robots and Systems*, pages 5000–5007, 2019.

[21] K. M. Wurm, R. Kümmerle, C. Stachniss, and W. Burgard. Improving robot navigation in structured outdoor environments by identifying vegetation from laser data. In *Proceedings of the IEEE/RSJ International Conference on Intelligent Robots and Systems*, pages 1217–1222, 2009.

[22] K.M. Wurm, H. Kretzschmar, R. Kümmerle, C. Stachniss, and W. Burgard. Identifying vegetation from laser data in structured outdoor environments. *Robotics and Autonomous Systems*, 62(5):675–684, 2014.